\documentclass[10pt,twocolumn,letterpaper]{article}

\usepackage{btas}
\usepackage{times}
\usepackage{epsfig}
\usepackage{graphicx}
\usepackage{array} %for putting figure in array
\usepackage{amsmath}
\usepackage{amssymb}
\usepackage{tabularx}
\usepackage{capt-of}
\usepackage{tabularx}
\usepackage{algorithm2e}
\usepackage{multirow}
\usepackage{subcaption}
\usepackage{float}
\usepackage{parskip}
\usepackage{enumitem}
\usepackage{bm}

\newenvironment{Table}
  {\par\bigskip\noindent\minipage{\columnwidth}\centering}
  {\endminipage\par\bigskip}
\newcommand{\mbf}[1]{\ensuremath{\mathbf{#1}}}

\newcommand*\tageq{\refstepcounter{equation}\tag{\theequation}}

\btasfinalcopy % *** Uncomment this line for the final submission

\ifbtasfinal\fi
\begin{document}

%%%%%%%%% TITLE
\title{Triplet Probabilistic Embedding for Face Verification and Clustering}
\author{Swami Sankaranarayanan \and Azadeh Alavi \and Carlos D.Castillo \and Rama Chellappa \vspace{1mm} \\ 
Center for Automation Research, UMIACS, University of Maryland, College Park, MD 20742\\
{\tt\small \{swamiviv,azadeh,carlos,rama\}@umiacs.umd.edu}
}
\maketitle
\thispagestyle{empty}
%%%%%%%%% ABSTRACT
\begin{abstract}
   Despite significant progress made over the past twenty five years,
   unconstrained face verification remains a challenging problem. This
   paper proposes an approach that couples a deep CNN-based approach with a
   low-dimensional discriminative embedding step, learned using triplet probability
   constraints to address the unconstrained face verification problem. Aside from
   yielding performance improvements, this embedding provides significant
   advantages in terms of memory and for post-processing operations like subject
   specific clustering. Experiments on the challenging IJB-A dataset show that
   the proposed algorithm performs close to the state of the
   art methods in verification and identification metrics, while requiring much
   less training data and training/test time. The superior performance of the
   proposed method on the CFP dataset shows that the representation learned by
   our deep CNN is robust to large pose variation. Furthermore, we demonstrate
   the robustness of deep features to challenges including age, pose, blur
   and clutter by performing simple clustering experiments on both IJB-A and LFW
   datasets.
\end{abstract}

\section{Introduction}
\label{sec:intro}
Recently, with the advent of curated face datasets like Labeled faces in the
Wild (LFW) \cite{lfw} and advances in learning algorithms like Deep neural nets,
there is more hope that the unconstrained face verification problem can be
solved. A face verification algorithm compares two given templates that are
typically not seen during training. Research in face verification has progressed
well over the past few years, resulting in the saturation of performance on the
LFW dataset, yet the problem of unconstrained face verification remains a
challenge. This is evident by the performance of traditional algorithms on the
publicly available IJB-A dataset (\cite{ijba15}, \cite{fvff15}) that was
released recently. Moreover, despite the superb performance of CNN-based
approaches compared to traditional methods, a drawback of such methods is the
long training time needed. In this work, we present a Deep CNN (DCNN)
architecture that ensures faster training, and investigate how much the
performance can be improved if we are provided domain specific data.
Specifically, our contributions are as follows:
\begin{itemize}
\item We propose a deep network architecture and a training scheme that ensures
  faster training time.
\item We formulate a triplet probability embedding learning method to improve
  the performance of deep features for face verification and subject clustering.
\end{itemize}

%\begin{figure}
%\includegraphics[width=0.5\textwidth]{pipeline.png}
%\caption{Training and Deployment processing pipeline}
%\label{fig:pipeline}
%\end{figure}

During training, we use a publicly available face dataset to train our deep
architecture. Each image is pre-processed and aligned to a canonical view before
passing it to the deep network whose features are used to represent the image. In the case of IJB-A dataset, the data is divided into 10 splits, each split containing a training set and a test set. Hence, to further improve performance, we learn the proposed triplet probability embedding using the training set provided with each split over the features extracted from our DCNN model. During the deployment phase, given a face template, we extract the deep features using the raw CNN model after implementing automatic pre-processing steps such as face detection and fiducial extraction. The deep features are projected onto
a low-dimensional space using the embedding matrix learned during training (note
that the projection involves only matrix multiplication). We use the
128-dimensional feature as the final representation of the given face template. 

This paper is organized as follows: Section \ref{SOA} places our work among the
recently proposed approaches for face verification. Section \ref{net} details
the network architecture and the training scheme. The triplet probabilistic
embedding learning method is described in Section \ref{tpe} followed by results
on IJB-A and CFP datasets and a brief discussion in Section
\ref{sec:results}. In Section \ref{sec:cluster}, we demonstrate the ability of
the proposed method to cluster a media collection from LFW and IJB-A
datasets.

\section{Related Work}\label{SOA}
%This work broadly consists of two components: the deep network used as a feature extractor and the learning procedure that projects the input features onto a discriminative low-dimensional space.
In the past few years, there have been numerous works in using deep features for tasks related to face verification. The DeepFace \cite{deepface14} approach uses a carefully crafted 3D alignment procedure to preprocess face images and feeds them to a deep network that is trained using a large training set. More recently, Facenet \cite{facenet15} uses a large private dataset to train several deep network models using a triplet distance loss function. The training time for this network is of the order of few weeks. Since the release of the IJB-A dataset \cite{ijba15}, there have been several works that have published verification results for this dataset. Previous approaches presented in \cite{wang15} and \cite{parkhi15} train deep networks using the CASIA-WebFace dataset \cite{casia14} and the VGG-Face dataset respectively, requiring substantial training time. This paper proposes a network architecture and a training scheme that needs shorter training time and a small query time. \\

The idea of learning a compact and discriminative representation has been around for decades. Weinberger \emph{et al.}  \cite{lmnn05} used a Semi Definite Programming (SDP)-based formulation to learn a metric satisfying pairwise and triplet distance constraints in a large margin framework. More recently, this idea has been successfully applied to face verification by integrating the loss function within the deep network architecture (\cite{facenet15}, \cite{parkhi15}). Joint Bayesian metric learning is also another popular metric used for face verification (\cite{fvf13},\cite{chen15wacv}). These methods either require a large dataset for convergence or learn a metric directly and therefore are not amenable to subsequent operations like discriminative clustering or hashing. Classic methods like t-SNE \cite{tsne}, t-STE \cite{tste} and Crowd Kernel Learning (CKL) \cite{CKL} perform extremely well when used to visualize or cluster a given data collection. They either operate on the data matrix directly or the distance matrix generated from data by generating a large set of pairwise or triplet constraints. While these methods perform very well on a given set of data points, they do not generalize to out-of-sample data. In the current work, we aim to generalize such formulations, to a more traditional classification setting, where domain specific training and testing data is provided. We formulate an optimization problem based on triplet probabilities that performs dimensionality reduction aside from improving the discriminative ability of the test data. The embedding scheme described in this work is a more general framework that can be applied to any setting where labeled training data is available.

%The objective of the optimization algorithm is to minimize the violations in the constraint set.
\section{Network Architecture}\label{net}

This section details the architecture and training algorithm for the deep
network used in our work. Our architecture consists of 7 convolutional layers
with varying kernel sizes. The initial layers have a larger size rapidly
subsampling the image and reducing the parameters while subsequent layers consist
of small filter sizes, which has proved to be very useful in face recognition
tasks (\cite{parkhi15},\cite{casia14}). Furthermore, we use the Parametric
Rectifier Linear units (PReLUs) instead of ReLUs, since they allow a negative
value for the output based on a learned threshold and have been shown to improve
the convergence rate \cite{prelu15}.

\begin{Table}

\centering
\resizebox{6cm}{2.2cm}{ 
\begin{tabular}{|c|c|c|}
% \multicolumn{3} \\ 
\hline
 Layer & Kernel Size/Stride & \#params\\
 \hline
 conv1    &11x11/4 & 35K\\
  pool1    &3x3/2 &    \\
 conv2    &5x5/2 & 614K\\
  pool2    &3x3/2 &    \\
 conv3    &3x3/1 & 885K\\
 conv4    &3x3/1 & 1.3M\\
 conv5    &3x3/1 & 2.3M\\
 conv6    &3x3/1 & 2.3M\\
 conv7    &3x3/1 & 2.3M\\
 pool7    &3x3/2 &    \\
 fc6    &1024 & 18.8M \\
 fc7    &512 & 524K \\
 fc8    &10548 & 10.8M \\
 Softmax Loss & &  Total: 39.8M\\
 \hline

\end{tabular}
}
\captionof{table}{Deep Network architecture details}
\label{arch}
\end{Table}

The top three convolutional layers (conv1-conv3) are initialized with the
weights from the AlexNet model \cite{alexnet12} trained on the ImageNet
challenge dataset. Several recent works (\cite{transfer1},\cite{transfer2}) have
empirically shown that this transfer of knowledge across different networks,
albeit for a different objective, improves performance and more significantly
reduces the need to train over a large number of iterations. 

The compared methods either learn their deep models from scratch
(\cite{parkhi15},\cite{nan}) or finetune only the last layer of fully
pre-trained models. The former results in large training time and the
latter does not generalize well to the task at hand (face verification) and
hence resulting in sub optimal performance. In the current work, even though we
use a pre-trained model (AlexNet) to initialize the proposed deep network, we do
so only for the first three convolutional layers, since they retain more generic
information (\cite{transfer1}). Subsequent layers learn representations which
are more specific to the task at hand. Thus, to learn more task specific
information, we add 4 convolutional layers each consisting of 512 kernels of
size $3\times3$. The layers conv4-conv7 do not downsample the input thereby
learning more complex higher dimensional representations. This hybrid
architecture proves to be extremely effective as our raw CNN representation
outperforms some very deep CNN models on the IJB-A dataset (Table 2 in Results).
In addition, we achieve that performance by training the proposed deep network using the
relatively smaller CASIA-WebFace dataset. 

The architecture of our network is shown in Table \ref{arch}. Layers
conv4-conv7 and the fully connected layers \textit{fc6-fc8} are initialized from
scratch using random Gaussian distributions. PReLU activation functions are added
between each layer. Since the network is used as a feature extractor, the last
layer \textit{fc8} is removed during deployment, thus reducing the number of
parameters to 29M. The inputs to the network are 227x227x3 RGB images. When the
network is deployed, the features are extracted from the \textit{fc7} layer
resulting in a dimensionality of 512. The network is trained using the Softmax
loss function for multiclass classification using the Caffe deep learning
platform \cite{caffe}.

\section{Learning a Discriminative Embedding}\label{tpe}
In this section, we describe our algorithm for learning a low-dimensional
embedding such that the resulting projections are more discriminative. Aside
from an improved performance, this embedding provides significant advantages in
terms of memory and enables post-processing operations like visualization and clustering. \\

Consider a triplet $t:=(\bm{v}_i,\bm{v}_j,\bm{v}_k)$, where $\bm{v}_i$ (anchor)
and $\bm{v}_j$ (positive) are from the same class, but $\bm{v}_k$ (negative)
belongs to a different class. Consider a function $S_W:\mathbb{R}^N \times
\mathbb{R}^N \mapsto \mathbb{R}$ that is parameterized by the matrix $\mbf{W}
\in \mathbb{R}^{n \times N}$, that measures the similarity between two vectors
$\bm{v}_i,\bm{v}_j \in \mathbb{R}^N$. Ideally, for all triplets $t$ that exist
in the training set, we would like the following constraint to be satisfied: 

\vspace{-5mm}
\begin{align*}
  S_\mbf{W}(\bm{v}_i,\bm{v}_j)> S_\mbf{W}(\bm{v}_i,\bm{v}_k)
\tageq \label{eq:constraint}
\end{align*}
Thus, the probability of a given triplet $t$ satisfying (\ref{eq:constraint}) can be written as:
\vspace{-2.5mm}
\begin{align*}
  p_{ijk}=\frac{e^{S_\mbf{W}(\bm{v}_i,\bm{v}_j)}}{e^{S_\mbf{W}(\bm{v}_i,\bm{v}_j)} + e^{S_\mbf{W}(\bm{v}_i,\bm{v}_k)}}
\tageq \label{eq:prob}
\end{align*}

The specific form of the similarity function is given as:
$S_\mbf{W}(\bm{v}_i,\bm{v}_j)= (\mbf{W}\bm{v}_i)^T \cdot (\mbf{W} \bm{v}_j)$. In our case, $\bm{v}_i$ and $\bm{v}_j$ are deep features normalized to unit length. To learn the embedding $\mbf{W}$ from a given set of triplets $\mathbb{T}$, we solve the following optimization:
\vspace{-2.5mm}
\begin{align*}
  \underset{\mbf{W}}{\text{argmin}} \sum_{ (\bm{v}_i,\bm{v}_j,\bm{v}_k) \in \mathbb{T}} - \log(p_{ijk}) 
\tageq \label{eq:train1}
\end{align*}

(\ref{eq:train1}) can be interpreted as maximizing the likelihood
(\ref{eq:constraint}) or minimizing the negative log-likelihood (NLL) over the
triplet set $\mathbb{T}$.  In practice, the above problem is solved in a
Large-Margin framework using Stochastic Gradient Descent (SGD) and the triplets
are sampled online. The gradient update for $\mbf{W}$ is given as:
\begin{align*}
  \mbf{W}_{\tau+1} = \mbf{W}_\tau - \eta * \mbf{W}_\tau *(1-p_{ijk})* (\bm{v}_i(\bm{v}_j-\bm{v}_k)^T \\
  + (\bm{v}_j-\bm{v}_k)\bm{v}_i^T)
\tageq \label{eq:update}
\end{align*}

where $\mbf{W}_\tau$ is the estimate at iteration $\tau$, $\mbf{W}_{\tau+1}$ is the
updated estimate, $(\bm{v}_i,\bm{v}_j,\bm{v}_k)$ is the triplet sampled at the current iteration and $\eta$ is the learning rate.\\ 

By choosing the dimension of $\mbf{W}$ as $n \times N$ with $n < N$, we achieve
dimensionality reduction in addition to improved performance. For our work, we fix
$n=128$ based on cross validation and $N=512$ is the dimensionality of our deep
features. $\mbf{W}$ is initialized with the first $n$ principal components of
the training data. At each iteration, a random anchor and a random positive data
point are chosen. To choose the negative, we perform hard negative mining, ie.
we choose the data point that has the least likelihood (\ref{eq:prob}) among the
randomly chosen 2000 negative instances at each iteration.

Since we compute the embedding matrix $\mbf{W}$ by optimizing over triplet
probabilities, we call this method Triplet Probability Embedding (TPE). The
technique closest to the one presented in this section, which is used in recent
works (\cite{facenet15},\cite{parkhi15}) computes the embedding \mbf{W} based on
satisfying a hinge loss constraint:
\begin{align*}
  \underset{\mbf{W}}{\text{argmin}} \sum_{ (\bm{v}_i,\bm{v}_j,\bm{v}_k) \in \mathbb{T}} \max\{0,\alpha + (\bm{v}_i-\bm{v}_j)^T\mbf{W^TW}(\bm{v}_i-\bm{v}_j)- \\
  (\bm{v}_i-\bm{v}_k)^T\mbf{W^TW}(\bm{v}_i-\bm{v}_k) \}
\tageq \label{eq:dist}
\end{align*}

\begin{figure}
\centering
\includegraphics[width=0.5\textwidth,height=0.30\textwidth]{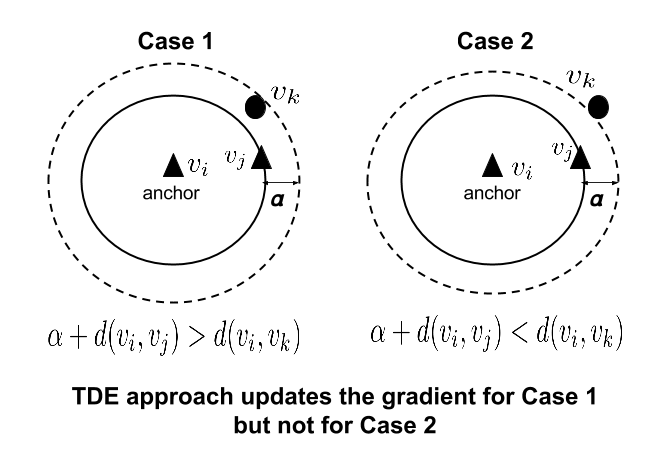}
\caption{Gradient update scenarios for the TDE method (\ref{eq:dist}). The notation is explained in the text}
\label{fig:tde}
\end{figure}

$\alpha$ acts a margin parameter for the loss function. To be consistent with
the terminology used in this paper, we call it Triplet Distance Embedding (TDE). To appreciate the difference between the two approaches, Figure \ref{fig:tde} shows the case where the gradient update for the TDE method
(\ref{eq:dist}) occurs. If the value of $\alpha$ is not appropriately chosen, a triplet is considered good even if the positive and negative are very
close to one another. But under the proposed formulation, both cases referred to
in Figure \ref{fig:tde} will update the gradient but their contribution to the
gradient will be modulated by the probability with which they violate the
constraint in  (\ref{eq:constraint}). This modulation factor is specified by the
$(1-p_{ijk})$ term in the gradient update for TPE in (\ref{eq:update}) implying
that if the likelihood of a sampled triplet satisfying (\ref{eq:constraint}) is
high, then the gradient update is given a lower weight and vice-versa. Thus, in
our method, the margin parameter ($\alpha$) is automatically set based on the
likelihood. \\

\begin{figure}
\includegraphics[width=0.5\textwidth,height=0.25\textwidth]{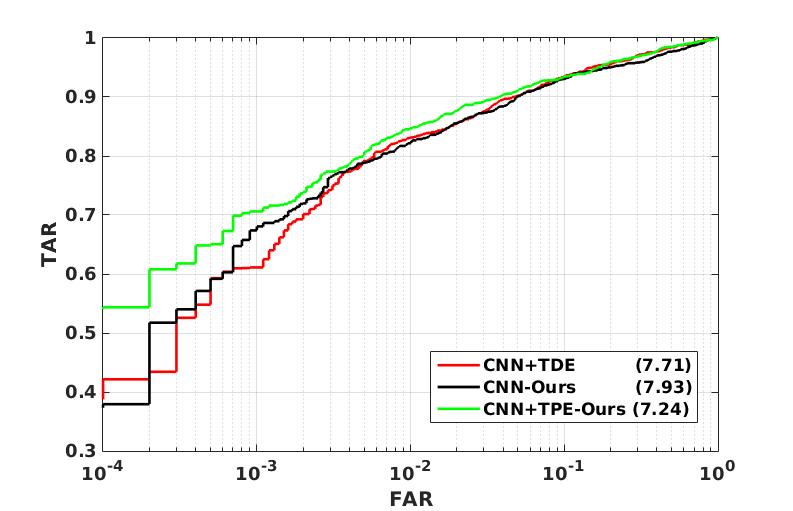}
\caption{Performance improvement on IJB-A split 1: FAR (vs) TAR plot. EER values are specified in brackets.}
\label{fig:tl}
\end{figure}

To compare the relative performances of the raw features before projection, with
TDE and with TPE (proposed method), we plot the traditional ROC curve (TAR (vs)
FAR) for split 1 of the IJB-A verify protocol for the three methods in Figure
\ref{fig:tl}. The Equal Error Rate (EER) metric is specified for each method. The performance
improvement due to TPE is significant, especially at regions of
FAR$=\{10^{-4},10^{-3}\}$. We observed a similar behaviour for all the ten
splits of the IJB-A dataset. 

\section{Experimental setup and Results}\label{sec:results}

\begin{figure}
\begin{subfigure}{.5\textwidth}
\centering
\includegraphics[width=0.6\linewidth]{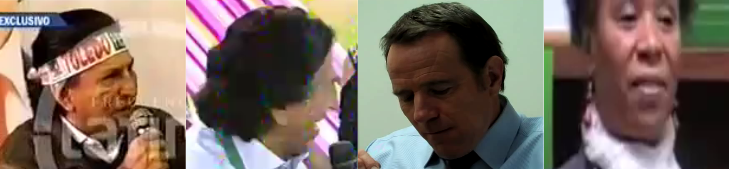}
\end{subfigure}
\caption{Images from the IJB-A dataset}
\label{fig:ijba-sample}
\end{figure}

\begin{figure}
\centering
\begin{subfigure}{.25\textwidth}
  \centering
  \includegraphics[width=.4\linewidth]{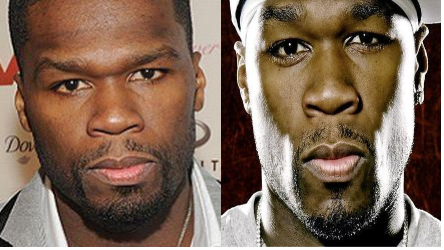}
  \caption{Frontal-Frontal}
  \label{fig:sub1}
\end{subfigure}%
\begin{subfigure}{.25\textwidth}
  \centering
  \includegraphics[width=.4\linewidth]{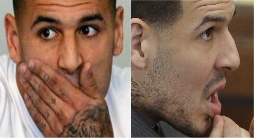}
  \caption{Frontal-Profile}
  \label{fig:sub2}
\end{subfigure}
\caption{Sample comparison pairs from the CFP dataset}
\label{fig:cfp}
\end{figure}

In this section we evaluate the proposed method on two challenging datasets: 
\begin{enumerate}[leftmargin=*]
\item \textbf{IARPA Janus Benchmark-A (IJB-A)}  \textbf{\cite{ijba15}}: This dataset contains 500
  subjects with a total of 25,813 images (5,399 still images and 20,414 video
  frames sampled at a rate of 1 in 60). The faces in the IJB-A dataset contain
  extreme poses and illuminations, more challenging than LFW \cite{lfw}.  Some sample images from the IJB-A dataset are shown in Figure \ref{fig:ijba-sample}. An additional challenge of the IJB-A verification protocol is that the template comparisons include image to image, image to set and set to set comparisons. In this work, for a given test template of the IJB-A data we perform two kinds of pooling to produce its final representation:

\begin{itemize}[leftmargin=*]
\item  \textit{Average pooling} (CNN$_{ave}$): The deep features of the images and/or
  frames present in the template are combined by taking a componentwise average
  to produce one feature vector. Thus each feature equally contributes to the
  final representation. 
\item  \textit{Media pooling} (CNN$_{media}$): The deep features are combined keeping in
  mind the media source they come from. The metadata provided with IJB-A gives
  us the \textit{media id} for each item of the template. Thus to get the final
  feature vector, we first take an intra-media average and then combine these by
  taking the inter-media average. Thus each feature's  contribution to the final
  representation is weighted based on its source.
\end{itemize}

 \item \textbf{Celebrities in Frontal-Profile (CFP) \cite{cfpw}}: This dataset
  contains 7000 images of 500 subjects. The dataset is used for evaluating how
  face verification approaches handle pose variation. Hence, it consists of 5000
  images in frontal view and 2000 images in extreme profile. The data is
  organized into 10 splits, each containing equal number of frontal-frontal and
  frontal-profile comparisons. Sample comparison pairs of the CFP dataset are
  shown in Figure \ref{fig:cfp}.  
\end{enumerate}

\subsection{Pre-processing}
In the training phase, given an input image, we use the HyperFace method
\cite{hyperface} for face detection and fiducial point extraction. The HyperFace
detector automatically extracts many faces from a given image. For the IJB-A dataset, since most images contain more than one face, we use the bounding boxes provided along with
the dataset to select the person of interest from the list of automatic
detections. We select the detection that has the maximum area overlap with the
manually provided bounding box. In the IJB-A dataset, there are few images for
which the HyperFace detector cannot find the relevant face. For the missed
cases, we crop the face using the bounding box information provided with the
dataset and pass it to HyperFace to extract the fiducials. We use six fiducial
points (eyes and mouth corners) to align the detected image to a canonical view
using the similarity transform. For the CFP dataset, since the six keypoints cannot
be computed for profile faces we only use three keypoints on one side of the
face for aligning them.

% \multicolumn{8}{|c|}{IJB-A: Verification Results (FNMRs at FMRs)} \\

\begin{table*}
\centering
\resizebox{1.02\textwidth}{2cm}{
\begin{tabular}{|c|c||c||c||c||c|c|c|}
 \hline
 \multirow{3}{*}{Method} &
      \multicolumn{3}{c|}{IJB-A Verification (FNMR@FMR)} &       
      \multicolumn{4}{c|}{IJB-A Identification} \\
 \hline
  & 0.001 & 0.01 & 0.1 & FPIR=0.01 &  FPIR=0.1 & Rank=1 & Rank=10 \\
 \hline
  GOTS \cite{ijba15} & 0.8 (0.008) &  0.59 (0.014) & 0.37 (0.023) & 0.047 (0.02) & 0.235 (0.03) & 0.443 (0.02) & - \\
 VGG-Face \cite{parkhi15} & 0.396 (0.06) &  0.195 (0.03) & 0.063(0.01) & 0.46 (0.07) & 0.67 (0.03) & 0.913 (0.01) & \textbf{0.981 (0.005)} \\
 Masi \emph{et al.} \cite{isi3d} & 0.275 & 0.114 & - & - & - & 0.906 & 0.977 \\ 
 NAN \cite{nan} & 0.215 (0.03) & 0.103 (0.01) & 0.041 (0.005) & - & - & - & - \\
 Crosswhite \emph{et al.} \cite{str-temp} & \textbf{0.135 (0.02)} &  \textbf{0.06 (0.01)} & \textbf{0.017 (0.007)} & \textbf{0.774 (0.05)} & \textbf{0.882 (0.016)} & 0.928 (0.01) & \textbf{0.986 (0.003)} \\  
 CNN$_{ave}$ (Ours) & 0.287 (0.05) &  0.146 (0.01) & 0.051 (0.006) & 0.626 (0.06) & 0.795 (0.02) & 0.90 (0.01) & 0.974 (0.004) \\
 CNN$_{media}$ (Ours) & 0.234 (0.02) &  0.129 (0.01) & 0.048 (0.005) & 0.67 (0.05) & 0.82 (0.013) & 0.925 (0.01) & 0.978 (0.005) \\
  CNN$_{media}$+TPE (Ours) & 0.187 (0.02) &  0.10 (0.01) & 0.036 (0.005) & 0.753 (0.03) & 0.863 (0.014) & \textbf{0.932 (0.01)} & 0.977 (0.005) \\
 \hline
\end{tabular}
}
\captionof{table}{Identification and Verification results on the IJB-A dataset. For identification, the scores reported are TPIR values at the indicated points. The results are averages over 10 splits and the standard deviation is given in the brackets for methods which have reported them. $'-'$ implies that the result is not reported for that method. The best results are given in bold.}
\label{results}
\end{table*}

\begin{table*}
\centering
  \begin{tabular}{|l|l|l|l|l|l|l|}
    \hline
    \multirow{3}{*}{Algorithm} &
      \multicolumn{3}{c|}{Frontal-Frontal} &       
      \multicolumn{3}{c|}{Frontal-Profile} 
       \\
       \hline
    & Accuracy & EER & AUC & Accuracy & EER & AUC \\
    \hline
    Sengupta \emph{et al.} \cite{cfpw} &  96.40 (0.69) & 3.48 (0.67) & 99.43 (0.31) & 84.91 (1.82) & 14.97 (1.98) & 93.00 (1.55)\\
    \hline
    Human Accuracy & 96.24 (0.67) & 5.34 (1.79)& 98.19 (1.13) & \textbf{94.57 (1.10)} & \textbf{5.02 (1.07)} & \textbf{98.92 (0.46)}  \\
    \hline
    CNN (Ours) & \textbf{96.93 (0.61)} & \textbf{2.51 (0.81)} & \textbf{99.68 (0.16)} & 89.17 (2.35) & 8.85 (0.99) & 97.00 (0.53) \\
    \hline
  \end{tabular}
  
\captionof{table}{Results on the CFP dataset \cite{cfpw}. The numbers are
averaged over ten test splits and the numbers in brackets indicate standard
deviations of those runs. The best results are given in bold.}
\label{cfpw-results}
\end{table*}

\subsection{Parameters and training times}
The training of the proposed deep architecture is done using SGD with momentum,
which is set to 0.9 and the learning rate is set to 1e-3 and decreased uniformly
by a factor of 10 every 50K iterations. The weight decay is set to 5e-4 for all
layers. The training batch size is set to 256. The training time for our deep
network is 24 hours on a single NVIDIA TitanX GPU. For the IJB-A dataset, we use
the training data provided with each split to obtain the triplet embedding which
takes 3 mins per split. This is the only additional splitwise processing that is
done by the proposed approach. During deployment, the average enrollment time per
image after pre-processing, including alignment and feature extraction is 8ms. 

\subsection{Evaluation Pipeline}
Given an image, we pre-process it as described in Section 5.1.  The deep
features are computed as an average of the image and its flip. Given two deep features to compare, we compute their cosine similarity score. More specifically,
for the IJB-A dataset, given a template containing multiple faces, we
\textit{flatten} the template features by average pooling or media pooling to
obtain a vector representation. For each split, we learn the TPE projection
using the provided training data. Given two templates for comparison, we compute the cosine similarity
score using the projected 128-dimensional representations. 

%The final representation is obtained as:
%$y=\mbf{W}x$, where $x$ is the deep feature and $\mbf{W}$ is the TPE projection
matrix.
\subsection{Evaluation Metrics}
We report two types of results for the IJB-A dataset: Verification and
Identification. For the verification protocol, we report the False Non-Match
Rate (FNMR) values at several False Match Rates (FMR). For the identification
results, we report open set and closed set metrics. For the open set metrics, the True Positive
Identification Rate quantifies the fraction of subjects that are classified
correctly among the ones that exist in probe but not in gallery. For the closed
set metrics, we report the CMC numbers at different values of False Positive
Identification Rates (FPIRs) and Ranks. More details on the evaluation
metrics for the IJB-A protocol can be found in \cite{ijba15}. 

For the CFP dataset, following the protocol set in \cite{cfpw}, we report the
Area under the curve (AUC) and Equal Error Rate (EER) values as averages across
splits, in addition to the classification accuracy. To obtain the accuracy for
each split, we threshold our CNN similarity scores where the threshold is set to
the value that provides the highest classification accuracy over the training data
for each split.

\subsection{Discussion}
\subsubsection*{Performance on IJB-A}
Table \ref{results} presents the results for the proposed methods compared to
existing results for the IJB-A Verification and Identification protocol. The
compared methods are described below:
\begin{itemize}[leftmargin=*]
\item Government-of-the-Shelf (GOTS) \cite{ijba15} is the baseline performance
  provided along with the IJB-A dataset.
\item Parkhi \emph{et al.} \cite{parkhi15} train a very deep network (22 layers) over the VGG-Face dataset which contains 2.6M images from 2622 subjects.
\item The Neural Aggregation network (NAN) \cite{nan} is trained over large
  amount of videos from the CELEB-1000 dataset \cite{celeb1000} starting from the GoogleNet \cite{gnet} architecture.
\item Masi \emph{et al.} \cite{isi3d} use a deep CNN based approach that includes a combination of in-plane aligned images, 3D rendered images to augment their performance. The 3D rendered images are also generated during test time per template comparison. It should be noted that many test images of the IJB-A dataset contain extreme poses, harsh illumination conditions and significant blur. 
\item Crosswhite \emph{et al.} use template adaptation \cite{wolf} to tune the performance of their raw features specifically to the IJB-A dataset.
\end{itemize} 
\vspace{2mm}

Compared to these methods, the proposed method trains a single CNN model on the
CASIA-WebFace dataset which consists of about 500K images and requires much
shorter training time and has a very fast query time (0.08s after face detection
per image pair). As shown in Table \ref{results}, our raw CNN features after media pooling perform better than most compared methods across both the  verification and identification protocols of the IJB-A dataset, with the exception of the template adaptation method by Crosswhite \emph{et al.} \cite{str-temp} which is discussed below. The TPE method provides significant improvement for both identification and verification tasks as shown in Table \ref{results}.\\

 The method by Crosswhite \emph{et al.} \cite{str-temp} uses the VGG-Face network \cite{parkhi15} descriptors (4096-d) as the raw features. They use the concept of template adaptation \cite{wolf} to improve their performance as follows: when pooling multiple faces of a given template, they train a linear SVM with the features of this template as positive and a
fixed set of negatives extracted from the training data of the IJB-A splits.
Let's denote the pooled template feature and classifier pair as $(t,w)$. Then, at
query time when comparing two templates $(t_1,w_1)$ and $(t_2,w_2)$, the similarity 
score is computed as: $\frac{1}{2} \left( t_1 \cdot w_2 + t_2 \cdot w_1
\right)$. Even when using a carefully engineered fast linear classifier training
algorithm, this procedure increases the run time of the pooling procedure. The query time per template comparison is also higher due to the high dimensionality of the input features. In contrast, the proposed approach requires a matrix multiplication and a vector dot product per 
comparison. By using a simple neural network architecture, a relatively smaller
training dataset and a fast embedding method we have realized a faster and more efficient end-to-end system. To improve our performance further, we are currently incorporating the use of video data into our approach. 
  
%\begin{table}
%\centering
%\begin{tabular}{l*{3}{c}r}
%\hline
%FMR & Ours &  Crosswhite \emph{et al.} \cite{str-temp} \\
%\hline 
%1e-2 & 0.10 (0.012) & 0.061 (0.013)  \\
% \hline
% 1e-1 & 0.036 (0.005) & 0.017 (0.007)  \\
% \hline
%\end{tabular}
%\caption{Recent results on the IJB-A verification protocol. Reported as FNMR at FMR}
%\label{tab:recent}
%\end{table}%

\subsubsection*{Performance on CFP}
On the CFP dataset, we achieve a new state-of-art on both Frontal-Frontal and
Frontal-Profile comparisons, the latter by a large margin. More specifically,
for the Frontal-Profile case, we manage to reduce the error rate by
\textbf{40.8\%}. It should be noted that for a fair comparison we have used our
raw CNN features without performing TPE. This shows that the raw CNN features we
learn are effective even at extreme pose variations.

\section{Clustering Faces}\label{sec:cluster}

\begin{figure*}
\begin{subfigure}{.5\textwidth}
  \centering
  \includegraphics[width=.6\linewidth]{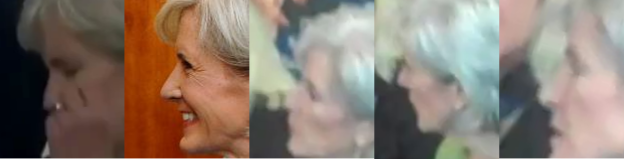}
  \caption{}
  \label{fig:sfig1}
\end{subfigure}%
\begin{subfigure}{.5\textwidth}
  \centering
  \includegraphics[width=.6\linewidth]{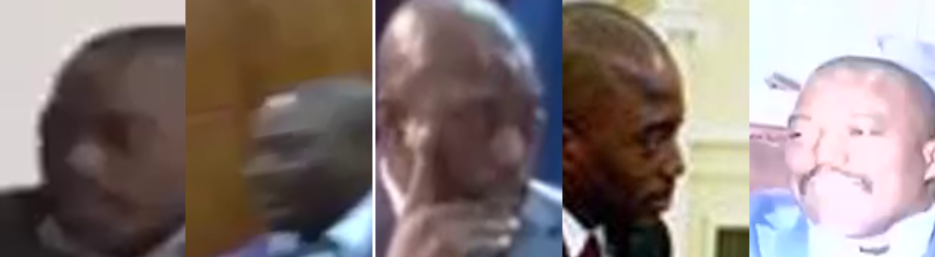}
    \caption{}
  \label{fig:sfig2}
\end{subfigure}\\
\begin{subfigure}{.5\textwidth}
  \centering
  \includegraphics[width=.6\linewidth]{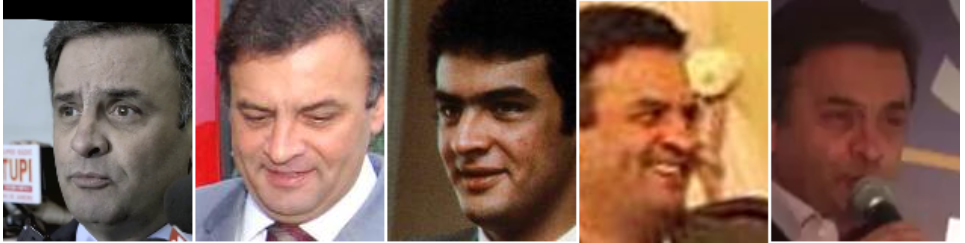}
    \caption{}
  \label{fig:sfig3}
\end{subfigure}%
\begin{subfigure}{.5\textwidth}
  \centering
  \includegraphics[width=.6\linewidth]{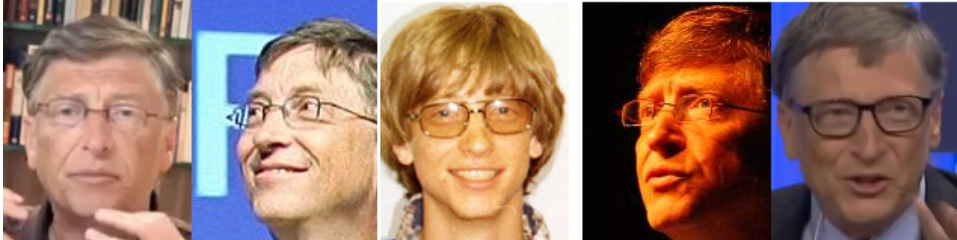}
    \caption{}
  \label{fig:sfig4}
\end{subfigure} \\
%\begin{subfigure}{.5\textwidth}
%  \centering
%  \includegraphics[width=.6\linewidth]{image_clusters/other/1.png}
%  \caption{}
%  \label{fig:sfig5}
%\end{subfigure}%
%\begin{subfigure}{.5\textwidth}
%  \centering
%  \includegraphics[width=.6\linewidth]{image_clusters/other/2.png}
%  \caption{}
%  \label{fig:sfig6}
%\end{subfigure}
\caption{Sample clusters output from the Clustering approach discussed in Section 6 for the data from the split 1 of the IJB-A dataset. Top row (a,b) shows robustness to pose and blur; Bottom row (c,d) contains clusters that are robust to age}
\label{fig:cluster}
\end{figure*}

This section illustrates how the proposed TPE method can be used to cluster a
given data collection. We perform two clustering experiments:
\begin{enumerate}[leftmargin=*]
\item We perform clustering on the entire LFW \cite{lfw} dataset that consists
  of 13233 images of 5749 subjects. It should be noted that about 4169 subjects have only one image.
\item We use the IJB-A dataset and cluster the templates corresponding to the query set for each split in the IJB-A verify protocol.
\end{enumerate}

For evaluating the clustering results, we use the metrics defined in
\cite{msu_cluster}. These are summarized below:
\begin{itemize}[leftmargin=*]
\item \textit{Pairwise Precision ($P_{pair}$)}: The fraction of pairs of samples
  within a cluster among all possible pairs which are of the same class, over
  the total number of same cluster pairs.
\item \textit{Pairwise Recall ($R_{pair}$)}: The fraction of pairs of samples
  within a class among all possible pairs which are placed in the same cluster,
  over the total number of same-class pairs.
\end{itemize}

Using these metrics, the F$_1$-score is computed as:
\begin{equation}
F_1 = \frac{2*P_{pair}*R_{pair}}{R_{pair}+P_{pair}}
\end{equation}

The simplest way we found to demonstrate the effectiveness of our deep features
and the proposed TPE method, is to use the standard MATLAB implementation of the
agglomerative clustering algorithm with the average linkage metric. We use the
cosine similarity as our basic clustering metric. The simple clustering
algorithm that we have used here has computational complexity of
$O(N^2)$. In its current form, this does not scale to large datasets with
millions of images. We are currently working on a more efficient and scalable (yet approximate) version of this algorithm.

\paragraph{Clustering LFW:-}The images in the LFW dataset are pre-processed as described in Section 5.1. For each image and its flip, the deep features are extracted using the proposed architecture, averaged and normalized to unit $L_2$ norm. We run the clustering algorithm over the entire data in a single shot. The clustering algorithm takes as input a cut-off parameter which acts as a distance threshold (below which any two clusters will not be merged). In our experiments, we vary this cut-off parameter over a small range and evaluate the
resulting clustering using the $F_1$-score. We pick the result that yields the
best $F_1$-score. Table \ref{tab:lfw_cluster} shows the result of our approach
and compares it to a recently released clustering approach based on approximate
Rank-order clustering \cite{msu_cluster}. It should be noted that, in the case
of \cite{msu_cluster}, the clustering result is chosen by varying the number of
clusters and picking the one with the best $F_1$-score. In our approach, we vary
the cut-off threshold which is the property of deep features and hence is a
more intuitive parameter to tune. We see from Table \ref{tab:lfw_cluster} that
aside from better performance, our total cluster estimate is closer to the
ground truth value of 5749 than \cite{msu_cluster}.

\begin{table}
\centering
\begin{tabular}{ | l | l | l|}
    \hline
    Method & $F_1$-score  & Clusters \\ \hline
    \cite{msu_cluster} & 0.87 & 6508 \\ \hline
    CNN (Ours) & \textbf{0.955} & 5351\\ \hline    
\end{tabular}
    \captionof{table}{$F_1$-score for comparison of the two clustering schemes on
    the LFW dataset. The ground truth cluster number is 5749.}
    \label{tab:lfw_cluster}
    \end{table}
\begin{table}
\centering
\resizebox{0.45\textwidth}{!}{
\begin{tabular}{ | l | l | l| l |}
    \hline
    Method & $F_1$-score & Clusters &  After Pruning \\ \hline
    CNN$_{media}$ & 0.79 (0.02) & 293 (22) & 173 \\ \hline
    CNN$_{media}$+TPE & \textbf{0.843 (0.03)} & 258 (17) & 167 \\ \hline    
\end{tabular}
}
    \captionof{table}{Clustering metrics over the IJB-A 1:1 protocol. The
    standard deviation is indicated in brackets. The ground truth subjects per
    each split is 167.}
\label{tab:ijba_cluster}
\end{table}

\paragraph{Clustering IJB-A:-}The IJB-A dataset is processed as described in Section 5. In this section, we aim to cluster the query templates provided with each split for the verify protocol. We report the results of two experiments: with the raw CNN features (CNN$_{media}$ in Table 2) and with the projected CNN features, where the projection matrix is learned through the proposed TPE method (CNN$_{media}$+TPE in Table 2). The cut-off threshold required for our clustering algorithm is learned automatically based on the training data, i.e. we choose the threshold that gives the maximum $F_1$-score over the training data. The scores reported in Table \ref{tab:ijba_cluster} are average values over ten splits. As expected, the TPE method improves the clustering performance of raw features. The subject estimate is the number of clusters produced as a direct result of our clustering algorithm. The pruned estimate is obtained by ignoring clusters that have fewer than 3 images. 

\begin{figure}
  \centering
  \includegraphics[width=0.5\textwidth,height=0.3\textwidth]{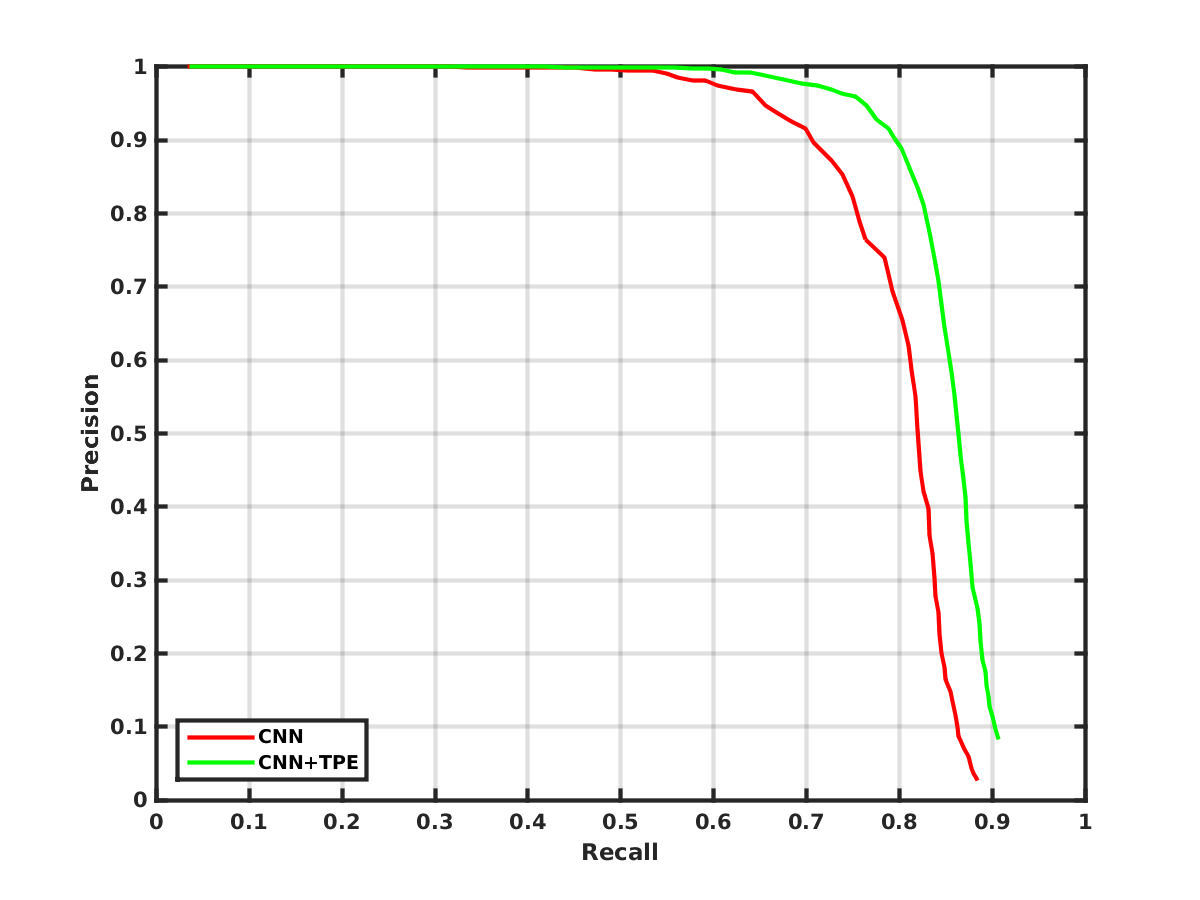}
  \caption{Precision-Recall curve plotted over cut-off threshold varied from 0 to 1.}
  \label{fig:pr_curve}
\end{figure}
 
For a more complete evaluation of our performance over varying threshold values, we plot the Precision-Recall (PR) curve for the IJB-A clustering experiment in Figure \ref{fig:pr_curve}. As can be observed, the PR curve for clustering the IJB-A data using embedded features exhibits a better performance at all operating points. This is a more transparent evaluation than reporting only the $F_1$-score since the latter effectively fixes the operating point but the PR curve reveals the performance at all operating points. 
\section{Conclusion and Future Work}\label{conclusion}
In this paper, we proposed a deep CNN-based approach coupled with a
low-dimensional discriminative embedding learned using triplet probability
constraints in a large margin fashion. The proposed pipeline enables a faster
training time and improves face verification performance especially at low
FMRs. We demonstrated the effectiveness of the proposed method on two
challenging datasets: IJB-A and CFP and achieved performance close to
the state of the art while using a deep model which is more compact and trained
using a moderately sized dataset. We demonstrated the robustness of our features
using a simple clustering algorithm on the LFW and IJB-A datasets. For future
work, we plan to use videos directly during training and also embed our TPE
approach into training the deep network. We intend to scale our clustering
algorithm to handle large scale scenarios such as large impostor sets of the
order of millions. 

\section{Acknowledgement}\label{sec:ackno}
This research is based upon work supported by the Office of the Director of National Intelligence (ODNI), Intelligence Advanced Research Projects Activity (IARPA),via IARPA R\&D Contract No. 2014-14071600012.  The views and conclusions contained herein are those of the authors and should not be interpreted as necessarily representing the official policies or endorsements, either expressed or implied, of the ODNI, IARPA, or the U.S. Government. The U.S. Government is authorized to reproduce and distribute reprints for Governmental purposes notwithstanding any copyright annotation thereon.

\bibliographystyle{ieeetr}
\bibliography{v2/refs}

\end{document}